\documentclass[10pt, a4paper]{article}

\usepackage[review]{lrec-coling2024} 
\usepackage{booktabs}

\usepackage[misc,geometry]{ifsym}

\usepackage{hyperref}
\RequirePackage{multirow}
\RequirePackage{array}
\RequirePackage{booktabs}
\RequirePackage{natbib}
\RequirePackage{multibib}
\RequirePackage{graphicx}
\RequirePackage{tabularx}
\RequirePackage{soul}
\RequirePackage{makecell}

\usepackage{tikz}
\usetikzlibrary{3d}
\usetikzlibrary{mindmap}
\usepackage{xcolor}

\RequirePackage{graphics}
\RequirePackage{linguex}
\usepackage{natbib}

\title{Benchmarking Large Language Models for Persian: \\ A Preliminary Study Focusing on ChatGPT}

\name{Amirhossein Abaskohi\normalsize{\Letter}\large$^1$, Sara Baruni\normalsize{\Letter}\large$^2$, Mostafa Masoudi\normalsize{\Letter}\large$^2$, \\
{\bf \large Nesa Abbasi$^{2*}$\thanks{$^{*}$ Equal contribution}, Mohammad Hadi Babalou$^{2*}$, Ali Edalat$^{2*}$, Sepehr Kamahi\normalsize{\Letter}\large$^{2*}$,}\\ 
{\bf \large Samin Mahdizadeh Sani$^{2*}$, Nikoo Naghavian$^{2*}$, Danial Namazifard$^{2*}$,}\\ 
{\bf \large Pouya Sadeghi$^{2*}$, Yadollah Yaghoobzadeh\normalsize{\Letter}\large$^{2,3}$}
$^\ddagger$
\thanks{$^\ddagger$See \ref{sec:appendix:contributions} for detailed author contributions.}}

\address{
    $^1$Computer Science Department, University of British Columbia, Vancouver, Canada \\
    $^2$School of Electrical and Computer Engineering \\
 College of Engineering, University of Tehran, Tehran, Iran \\
    $^3$Tehran Institute for Advanced Studies, Khatam University, Tehran, Iran \\
    \Letter aabaskoh@cs.ubc.ca, sara.baruni@ut.ac.ir, \\
    \Letter mostafa.masoudi@ut.ac.ir, sepehr.kamahi@ut.ac.ir \\
    \Letter y.yaghoobzadeh@ut.ac.ir
}

\abstract{This paper explores the efficacy of large language models (LLMs) for Persian. 
While ChatGPT and consequent LLMs have shown remarkable performance in English, their efficiency for more low-resource languages remains an open question. 
We present  the first comprehensive benchmarking study of LLMs across diverse Persian language tasks.
Our primary focus is on GPT-3.5-turbo, but we also include GPT-4 and OpenChat-3.5 to provide a more holistic evaluation. Our assessment encompasses a diverse set of tasks categorized into classic, reasoning, and knowledge-based domains. To enable a thorough comparison, we evaluate LLMs against existing task-specific fine-tuned models.
Given the limited availability of Persian datasets for reasoning tasks, we introduce two new benchmarks: one based on elementary school math questions and another derived from the entrance exams for 7th and 10th grades.
Our findings reveal that while LLMs, especially GPT-4, excel in tasks requiring reasoning abilities and a broad understanding of general knowledge, they often lag behind smaller pre-trained models fine-tuned specifically for particular tasks. Additionally, we observe improved performance when test sets are translated to English before inputting them into GPT-3.5.
These results highlight the significant potential for enhancing LLM performance in the Persian language. This is particularly noteworthy due to the unique attributes of Persian, including its distinct alphabet and writing styles. We have made our codes, prompts, and data available here:  \url{https://github.com/Ipouyall/Benchmarking_ChatGPT_for_Persian}.
\\ 
\newline  }

\begin{document}

\maketitleabstract

\section{Introduction}
The recent surge in AI breakthroughs owes much to the remarkable achievements of foundation models (FMs) across a multitude of research domains and practical applications \citep{schneider2022foundation}. These FMs, with their staggering capacity, have marked a new era of AI capabilities. From machine translation to question-answering, automatic speech recognition to text-to-speech generation, FMs have established themselves as invaluable tools \cite{10.5555/3495724.3495883,min2023exploring, touvron2023llama}, pushing the boundaries of what AI can achieve. At the forefront of this AI revolution are large language models (LLMs), exemplified by the GPT (generative pre-trained transformer) series \citep{10.5555/3495724.3495883}, which have demonstrated their skill across a plethora of language-related tasks.

LLMs, such as GPT-3.5-turbo \citep{openai2023gpt35}, are the pinnacle of natural language understanding and generation. These models, built upon the powerful transformer architecture \citep{10.5555/3295222.3295349} and trained on colossal datasets, possess the unique ability to predict the next token in a sequence, thereby capturing intricate linguistic patterns and nuances. Furthermore, when subjected to multilingual training, they exhibit a remarkable aptitude for understanding and generating text across different languages, transcending linguistic boundaries \citep{lin-etal-2022-shot, scao2022bloom, zhu2023multilingual}.

\begin{figure}
    \centering
    \includegraphics[width=1\linewidth]{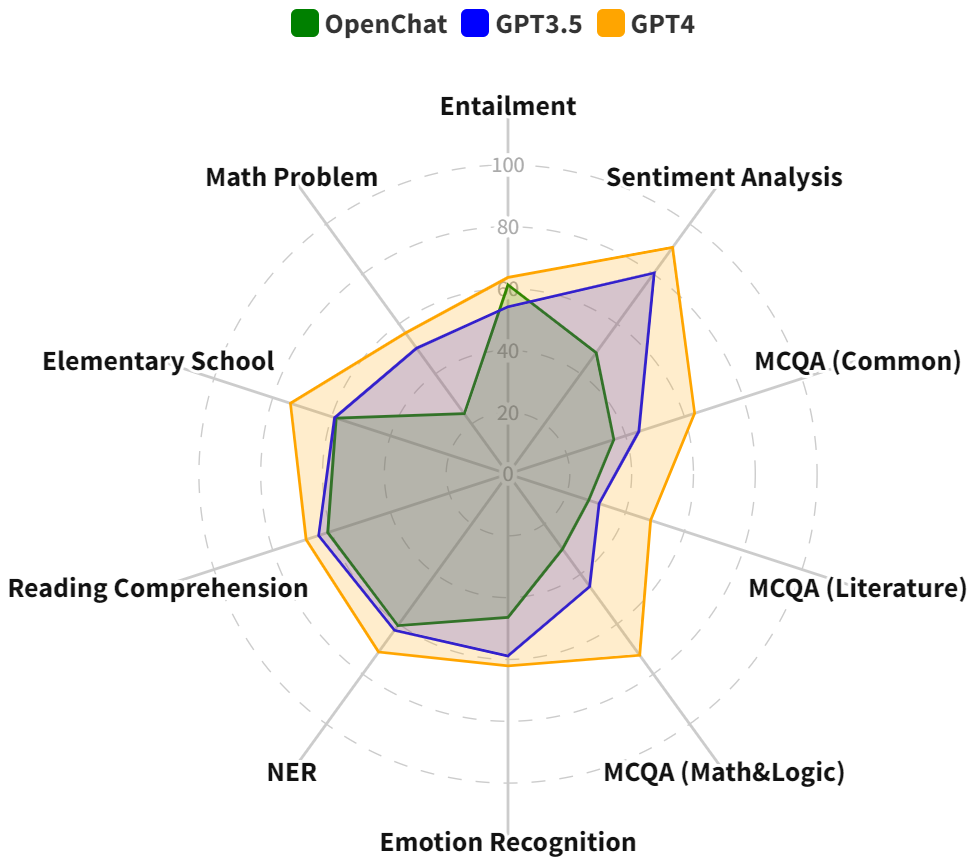}
    \caption{Performance of GPT-3.5, GPT-4, and OpenChat-3.5 models across various Persian NLP tasks. These are the best results of each model as detailed in Table \ref{tab:classictasks} and \ref{tab:openchat}.}
    \label{fig:results_overview}
\end{figure}

With the growing accessibility of models, there has been a growing interest in evaluating their performance on standard NLP tasks \citep{wu2023qualifying, khondaker2023gptaraeval, lai2023chatgpt}. 
Given the remarkable capabilities and impact of LLMs on downstream applications and human-AI interaction, it is crucial to evaluate their performance comprehensively and rigorously. Initiatives like the holistic evaluation of language models (HELM) \citep{liang2022holistic} and BIG-Bench \citep{srivastava2023beyond} have emerged to assess the capabilities of these models, using diverse metrics and scenarios. These evaluations enhance our understanding of the strengths and limitations of LLMs in various linguistic contexts.

In this paper, we focus on evaluating multilingual LLMs in the Persian language. Persian is an Indo-European language with a rich history and cultural significance. It is spoken by millions of people worldwide, mainly in Iran, Afghanistan, and Tajikistan. Persian presents a unique challenge and opportunity for benchmarking LLMs due to its linguistic features, such as its distinct character set compared to many high-resource languages (especially those with Latin characters), and its multiple writing styles and dialects \citep{shamsfard2019challenges}.

Our paper presents the results of an extensive benchmarking study focusing on evaluating the performance of ChatGPT, specifically GPT-3.5-turbo.
To ensure a comprehensive assessment of current LLMs, we also investigate GPT-4 and an open-source model called OpenChat-3.5 \citep{wang2024openchat}.
We design our benchmarking study based on several criteria and principles, such as the relevance, difficulty, diversity, and quality of the tasks and datasets. 
We include tasks such as sentiment analysis, question answering, natural language inference, and translation, with content derived from various sources, including articles and social media. 
We leverage publicly available datasets that have been curated or annotated for these tasks. We also create benchmarks ourselves for tasks that are not well-represented in the existing datasets or that require specific knowledge or skills from LLMs, such as elementary school and mathematical question answering. 

Throughout this benchmarking endeavor, we aim to address fundamental questions in the context of the Persian language:
(I) How do multilingual LLMs perform in the Persian language given only an instruction and few-shot examples for each task?
(II) How do the performance and effectiveness of prompting strategies vary across tasks of varying complexity, from sequence tagging to classification and reasoning, within the Persian NLP landscape?
(III) How do LLMs compare to readily available state-of-the-art models for these tasks in the Persian language?

We try to answer these questions through a preliminary evaluation, offering insights, findings, and unique contributions in the context of Persian NLP. 
In summary, our evaluation highlights the following key findings:
(I) ChatGPT's performance is lower compared to supervised models in classic tasks, but it outperforms them in QA reasoning and sentiment analysis tasks. 
(II) ChatGPT has a good grasp of common knowledge but lacks knowledge of Persian literature.
(III) Writing task descriptions in English outperforms Persian most tasks.
(IV) Unlike GPT-4, in most cases of GPT-3.5 experiments, increasing the number of shots does not lead to better results, except when it is used to understand the structure of the output, like in a reading comprehension task.
(V) Translating examples into English enhances GPT-3.5’s performance across most tasks.
(VI) As illustrated in Figure~\ref{fig:results_overview}, GPT-4 emerges as the most effective model, followed by GPT-3.5 and OpenChat-3.5, respectively, in terms of overall performance.
(VII) Remarkably, OpenChat-3.5—an open-source model with approximately seven billion parameters—demonstrates strong performance across several tasks when compared to both GPT-3.5 and GPT-4.


\section{Related Work}
\label{sec:related-work}
ChatGPT is a powerful LLM that has been applied to various tasks and domains in NLP and beyond. A comprehensive overview of ChatGPT-related research is provided by \citet{liu2023summary}, who discuss its potential applications and challenges. 
Evaluating LLMs is a crucial step in their development and improvement, but it is also a complex and multifaceted problem. \citet{chang2023survey} conduct a thorough survey of LLM evaluation methods, covering the questions of what, how, and where to evaluate LLMs. 
To assess the capabilities and limitations of ChatGPT and its successor GPT-4 \cite{openai2023gpt4}, \citet{mao2023gpteval} perform a systematic evaluation of their linguistic and reasoning skills, scientific knowledge, and ethical aspects.
In the following, we review some of the studies related to the evaluation of LLMs in multilingual and non-English settings.

\paragraph{Multilingual evaluation}

ChatGPT’s performance on thirty-seven languages and seven NLP tasks is evaluated by \citet{lai2023chatgpt}. They find that ChatGPT is inferior to mT5-XXL in all tasks and languages, especially in part-of-speech tagging, named entity recognition, relation extraction, and summarization. ChatGPT tends to generate long summaries, which may affect its performance negatively. ChatGPT also shows different strengths and weaknesses across tasks and languages, with better results in grammar-related tasks than in semantic-related tasks. ChatGPT sometimes exceeds expectations, outdoing high-resource languages in some low-resource language tasks. ChatGPT performs more consistently when prompted in English, showing better comprehension and analysis skills. We also examine this effect in our Persian benchmarks. \citet{bang2023multitask} also test ChatGPT on various tasks and highlight its limitations in language understanding and reasoning.

ChatGPT’s ability to answer multi-choice questions in the Iranian medical residency entrance exam is explored by \citet{khorshidi2023application}. They test ChatGPT on questions in the original language (Persian) and questions translated into English, French, and Spanish, and find that ChatGPT answers correctly more than 80\% of them in each language. They also note that ChatGPT can translate the questions without losing their originality. 
Here, we also do an analysis of translating Persian examples in different tasks into English, and how that affects the performance of ChatGPT. We employ Google Translate.

\paragraph{Monolingual evaluation}
\citet{khondaker2023gptaraeval} evaluate ChatGPT on thirty-two diverse Arabic natural language understanding (NLU) and natural language generation (NLG) tasks. They have observed that in a few-shot setting, ChatGPT surpasses the performance of significantly smaller specialized models fine-tuned specifically for Arabic.
Moreover, \citet{abdelali2023benchmarking} conduct a thorough assessment of LLMs in the context of Arabic NLP and speech processing tasks. Their study involves a comparison with SOTA task-specific models and an exploration of various prompt strategies. The findings indicate that while some tasks achieve performance levels on par with or exceeding task-specific models, the majority of tasks exhibit underperformance.

\citet{wu2023qualifying} assess ChatGPT's performance on the China national medical licensing examination (CNMLE) and introduce an innovative approach aimed at enhancing ChatGPT. This improvement involves the integration of medical domain knowledge and the facilitation of few-shot learning.
In the study by \citet{kasai2023evaluating}, GPT-4, GPT-3, and ChatGPT undergo evaluation in the context of Japanese medical licensing exams. The results of the evaluation demonstrate that GPT-4 surpasses both ChatGPT and GPT-3 in performance.

\paragraph{Machine translation}

\citet{stap-araabi-2023-chatgpt} investigate the challenges of machine translation (MT) systems for indigenous and extremely low-resource languages. The study evaluates the translation performance of Spanish to eleven indigenous languages from South America. The results show that even LLMs like GPT-4 are not well-suited for extremely low-resource languages. 
\citet{zhu2023multilingual} assess LLMs for multilingual MT. They evaluate some LLMs in 102 languages and compare their performance to a supervised baseline. The results show that even the best-performing LLM, ChatGPT, lags behind the baseline in the majority of translation directions.

\begin{table}
\centering
\resizebox{\columnwidth}{!}{%
\footnotesize
\begin{tabular}{ll}
\toprule
\textbf{Task} & \textbf{Dataset} \\
\toprule
Sentiment Classification & ParsiNLU \\
Reading Comprehension  & ParsiNLU  \\
Translation & ParsiNLU  \\
Emotion Classification & ArmanEmo  \\
NER & ArmanNER  \\
Textual Entailment  & ParsiNLU +  ConjNLI* \\
Multiple-choice QA & ParsiNLU \\
Elementary School Questions & Ours  \\
Mathematical Problems & Ours + MATH* \\
\bottomrule
\end{tabular}%
}
\caption{Our target tasks and the corresponding datasets. Translated datasets are specified by *.}
\label{dataset}
\end{table}

\section{Tasks and Datasets}

\label{sec:datasets}

We investigate the performance of GPT-3.5, GPT-4, OpenChat-3.5, and also SOTA models across a range of NLP tasks, grouped into classic, reasoning, and knowledge categories.
Table \ref{dataset} displays the benchmarks utilized for task evaluation. In the following sections, we present the task descriptions along with the corresponding benchmarks employed.

\subsection{Classic Tasks}

We categorize classic NLP tasks into three groups: classification, span retrieval, and generation. We describe each group and its tasks in detail below.

\subsubsection{Classification}
This category encompasses classic classification tasks, including sentiment analysis, emotion recognition, and named entity recognition. These foundational tasks are essential in NLP and hold a crucial role in various applications. Our examination of these classification tasks seeks to shed light on the performance and capabilities of the models under scrutiny in handling these essential tasks effectively.

The task of \textbf{sentiment analysis} aims to identify the emotional tone conveyed in a text, revealing the author’s feelings and opinions. We use the sentiment analysis part of the ParsiNLU dataset \citep{khashabi-etal-2021-parsinlu} for this task, focusing on the food domain. We classify examples into four categories: ``positive'', ``neutral'', ``negative'', and ``other''\footnote{The ``other'' label encapsulates sentiments that do not neatly fit into predefined categories or are ambiguous.}. We note that the original ParsiNLU dataset has two more categories, ``very positive'' and ``very negative'', which we have merged into the ``positive'' and ``negative'' labels. 

\textbf{Emotion recognition} is a crucial task in understanding human sentiment. We utilize the ArmanEmo dataset \citep{https://doi.org/10.48550/arxiv.2207.11808}, comprising 7,308 samples divided into training 6,157 and testing 1,151 sets. The dataset covers Paul Ekman's six fundamental emotions: ``Happiness'', ``Sadness'', ``Anger'' ``Fear'', ``Wonder'' and ``Hatred'' \citep{8f15e737-2dc5-321d-b763-876ad64e340b} along with a category for ``Other''. These samples are collected from diverse sources, including social media like Twitter and Instagram, as well as the Iranian e-commerce company Digikala. 

In \textbf{Named entity recognition}, the goal is to accurately identify named entity mentions in text, including names and locations. For that, we use the ArmanNER dataset \citep{poostchi-etal-2016-personer}. The dataset includes 7,682 Persian sentences. The labels include ``person'', ``location'', ``organization'', ``product'', ``event'', and ``facility''.

\subsubsection{Span Retrieval}
\textbf{reading comprehension}, requires finding an answer span in a paragraph for a given question. We use the reading comprehension part of the ParsiNLU dataset \citep{khashabi-etal-2021-parsinlu} for this evaluation. This dataset contains a variety of natural and common questions asked by Persian users, taken from Google’s auto-completion feature. The paragraphs and answers are annotated by native Persian speakers, ensuring the linguistic quality and relevance of the dataset for Persian language users.

\subsubsection{Generation}
 In this group, we tackle the challenging task of Persian$\leftrightarrow$English \textbf{MT}, which requires automatically translating text from Persian to English, and vice versa. This task is essential for facilitating cross-lingual communication and understanding. To measure the performance of our translation systems, we use the MIZAN \citep{kashefi2020mizan} dataset from ParsiNLU \citep{khashabi-etal-2021-parsinlu}, which consists of 1,021,596 sentence pairs obtained from a manual alignment of famous literary texts with their official Persian translations.

\subsection{Reasoning}
We divide our assessments into two main categories: ``Math \& Logic'' and ``Classification''.
This categorization allows us to examine LLM’s performance on tasks that require mathematical and logical reasoning as well as those that involve classification, providing a comprehensive view of its reasoning skills.

\subsubsection{Classification}
In this group, we choose \textbf{textual entailment}, which requires different reasoning skills. The purpose of this task is to determine whether ``hypothesis'' logically follows or can be inferred from ``premise''. 
We adopt two datasets for this task, one Persian dataset, ParsiNLU \citep{khashabi-etal-2021-parsinlu}, and another English dataset, ConjNLI \citep{saha-etal-2020-conjnli}. The textual entailment section of the Parsinlu dataset contains some natural Persian sentences from Persian Wikipedia, Miras\footnote{\url{https://github.com/miras-tech/MirasText}}, and 
VOA\footnote{\url{https://jon.dehdari.org/corpora}}.
It also includes an MNLI dataset translated into Persian by Google translator with the revision of the Persian native speakers.
On the other hand, for the second dataset, we selected the ConjNLI English dataset and then translated examples of it into Persian using Google Translate.
Like other known entailment datasets, the labels of these datasets are ``entailment'', ``contradiction'', and ``neutral''.

\subsubsection{Math \& Logic}

\textbf{Multiple-choice QA (Math \& Logic)} is designed to assess a model's ability to solve mathematical and logical questions by external reasoning. We use the math and logic questions of the multiple-choice part of the ParsiNLU dataset \citep{khashabi-etal-2021-parsinlu}. This dataset contains 677 Persian questions in the ``Math \& Logic'' domain.

\textbf{Elementary school QA} is a dataset of multiple-choice questions without any context. The dataset consists of 50 Persian questions with two options each, collected by us from elementary school math exams in Iran. These questions are taken from public elementary-level exam questions of Ghalamchi Institute\footnote{\url{https://www.kanoon.ir/Public/ExamQuestions}}.

The \textbf{Mathematical problems QA} dataset consists of questions from Iran’s National Organization for Development of Exceptional Talents (Sampad\footnote{\url{https://sampad.gov.ir/}}) entrance exams for 7th and 10th grades, and translations from the MATH dataset \citep{hendrycksmath2021}. We selected questions from Sampad exams that were clear and not too hard to compute. We also examined about 11,000 questions from the MATH dataset and chose those that were linguistically challenging but not too complex mathematically. We translated these questions into Farsi and standardized their format. We faced some LaTex-related issues that required LaTex coding skills to solve. To focus on evaluating the model’s mathematical understanding, we excluded questions with these LaTex problems. The final dataset has 179 samples, divided into five categories: Algebra, Counting and Probability, Sequences, Geometry, and Number Theory. We did not include questions in the Calculus category because they were usually too difficult to calculate. The distribution of our dataset across different categories is: Algebra (33.1\%), Counting and Probability (24.7\%), Sequences (20.2\%), Geometry (18.5\%), and Number Theory (3.4\%). This dataset is a useful benchmark for testing LLMs in zero- and few-shot settings, showing their ability to solve mathematical problems in different domains in Persian.

\subsection{Knowledge}

\textbf{Multiple-choice QA (literature and common-knowledge)} datasets are designed to assess the models' ability to answer questions using their encoded knowledge. A model should choose the right answer among the provided options given a question. We use the ``literature'' and ``common-knowledge'' domains of the multiple-choice part of the ParsiNLU dataset \citep{khashabi-etal-2021-parsinlu}, which include 834 and 949 questions, respectively.

\section{Prompt Engineering}
\label{sec:prompt-engineering}

A prompt consists of input queries or instructions that guide an LLM. Its format significantly influences the quality and relevance of the generated responses. A good prompt is essential for the model to understand user intent.
In this section, we elucidate the procedure involved in crafting the comprehensive template for our prompts. Subsequently, we delve into the various aspects that are subject to modification in the evaluation process.

\subsection{Prompt Structure}
 The overall process of crafting prompts in our work is as follows:
(I) First, we give a task description to tell the model what to do. This section may include both general descriptions and example language expressions.
(II) For classification tasks, we provide label names (e.g., sentiment analysis labels such as positive, negative, neutral, and other). Sometimes, we also give detailed explanations for each label (e.g., explaining textual entailment labels) and specify the expected response from the LLM model for easy post-processing (e.g., choosing from the given labels). (III) Subsequently, when necessary for certain tasks, we incorporate a template illustrating the expected format for both input and output in each example. Following this, we give sample examples for k-shot scenarios (where k > 0) within the template.
(IV) Finally, we insert test input in the prescribed format.

We also look into prior research endeavors, specifically their provided prompts in analogous studies such as 
\citet{khondaker2023gptaraeval} and
\citet{abdelali2023benchmarking} to refine our prompts.  Further, by employing ChatGPT's web interface, we try different prompts to ascertain the most effective ones in generating precise predictions while maintaining the desired output format.
It is noteworthy to mention that we have employed the chain-of-thought (CoT) prompting technique in the evaluation of the math problem task.

\subsection{Adjustable Elements}

We explore variations in the prompt language and the number of shots in our experimental setup.
In our main study, the task examples in k-shot and test inputs are in Persian\footnote{We also analyze the results of translating dataset samples into English in Section \ref{sec:sample_translation}}, but we test with the other sections of each prompt in both English and Persian. 
This lets us evaluate how the prompt language affects the model’s performance.
To elaborate, our experiments cover zero-shot, one-shot, and three-shot settings, utilizing both Persian and English prompts. It's noteworthy that in the k-shot mode for classification tasks, k samples are selected per class. 

An illustrative example of a prompt is presented in Table \ref{tab:prompt_example}.

\begin{table}[]
    \centering
    \small
\begin{tabular}{p{2.8in}}
    \\
    \toprule
    \textbf{Task Description:}
Natural Language Inference: Read the
following premise and hypothesis carefully
and determine the relationship between them.
Choose one of the three categories below that
best describes their
Relationship:    \\

       \textbf{Label Description:} \\
        - entailment: The meaning of the hypothesis is logically
inferred or derived from the premise.\\
- contradiction: The meaning of the hypothesis contradicts or
conflicts with the premise.\\
- neutral: There is no clear logical relationship
between the premise and hypothesis.
Note: The premise and hypothesis are in
Persian.\\
\textbf{Example Pattern:}\\
<premise><sep><hypothesis>\\
<category>:\\
entailment or contradiction or neutral\\
\\
\textit{<In this reaction, the speed of halogenation is independent of the concentration of halogen, but it depends on the concentration of ketone and acid.><sep><In this reaction, the speed of halogenation is dependent on the concentration of halogen, but it is independent of the concentration of ketone and acid.>} \\
<category>: \\
? \\
\bottomrule
    \end{tabular}
\caption{An example of how we construct our prompt for the NLI task.}
    \label{tab:prompt_example}
\end{table}




\section{Experiments and Results}
\label{sec:experiments}

In our main experiments, we assess the performance of the two prominent OpenAI models: ``gpt-3.5-turbo-0613''\citep{openai2023gpt35} and ``gpt-4'' \cite{openai2023gpt4}, utilizing a temperature setting of 0. Our assessment involves comparing these models with previously released state-of-the-art (SOTA) models for each task.
The SOTA models are models such as mT5, ParsBERT, and XLM-RoBERTa fine-tuned specifically for each task using the training sets provided in the datasets' reference papers, including ParsiNLU, ArmanEmo, and ArmanNER. The specific SOTA models for each task are listed in Appendix~\ref{appen:SOTA}.\footnote{Since we developed the Elementary School and Math Problems datasets for evaluation purposes internally, there is currently no training set, and therefore, there are no available SOTA reports for them.}

In our evaluation methodology, we randomly select 200 samples from the test sets provided in the ParsiNLU, ArmanEmo, and ArmanNER to form the test subset for each task. 
The decision to utilize a subset of 200 samples for each task aligns with the methodologies adopted in similar research, such as 
\citep{khondaker2023gptaraeval}. The constraint of LLMs' usage costs underscored the need for a selective approach. Our sample selection takes into account both the distribution of labels for the classification tasks and the randomization of data for a more comprehensive and unbiased representation.
Our comparisons with SOTA models are also done on the same 200 samples and therefore it is fair in that respect.

\begin{table*}[h!]
\centering
\resizebox{16.1cm}{!}
{\begin{tabular}{llccccccc|cccccc|cc}
\toprule
 & & &\multicolumn{6}{c}{GPT-3.5} & \multicolumn{6}{c}{GPT-4} & & \\
 \cmidrule(lr){4-15}
 & & & \multicolumn{3}{c}{\makecell{Persian Prompt \\ (N-shot)}} &\multicolumn{3}{c}{\makecell{English Prompt\\ (N-shot)}} & \multicolumn{3}{c}{\makecell{Persian Prompt \\ (N-shot)}} &\multicolumn{3}{c}{\makecell{English Prompt\\ (N-shot)}} & & \\
\cmidrule(lr){4-6}\cmidrule(lr){7-9}\cmidrule(lr){10-12}\cmidrule(lr){13-15}
  & Task & & 0 & 1 & 3 & 0 & 1 & 3 & 0 & 1 & 3 & 0 & 1 & 3 & SOTA & Random\\
  
\midrule
 \multirow{6}{*}{\rotatebox{90} {Classic}} & Sentiment analysis& Macro F1 & .725 & .804 & .791 & .786 & .798 & .761 & .786 & .825 & .856 & .812 & .878 & \textbf{.906} & .891 & .403 \\
& Emotion recognition & Macro F1 & .492 & .513 & .537 & .562 & .568 & .589 & .506 & .524 & .567 & .574 & .582 & .621 &\textbf{.699} & .117 \\
& NER & Macro F1 & .578 & .472 & .589 & .617 & \ .620 & .625 & .594 & .576 & .608 & .625 & .678 & .712 & \textbf{.988} & .041  \\
\cmidrule(lr){2-17}
 & MT ($En \rightarrow Fa$) & Bleu & 7.5 & 6.9 & 7.3 & 7.0 & 7.3 & 7.0 & 7.8 & 8.1 & 8.5 & 8.3 & 8.5 & \textbf{8.7} & 6.2 & - \\
 & MT ($Fa \rightarrow En$) & Bleu &10.5 & 10.8 & 11.0 & 11.0 & 11.0 & 10.8 & 10.8 & 11.0 & 11.2 & 11.0 & 11.2 & 11.5 & \textbf{11.7} & - \\
 \cmidrule(lr){2-17}

& Reading comprehension & F1 & .535 & .643 & .642 & .588 & .644 & .644 & .564 & .646 & .659 & .603 & .673 & .687 & \textbf{.691} & - \\

\midrule \midrule 

\multirow{5}{*}{\rotatebox{90}{Reasoning}} & {Textual entailment (Parsinlu)} & Macro F1 & .375 & .318 & .320 & .536 & .541 & .516 & .383 & .423 & .461 & .582 & .616 & .636 &\textbf{.690} & .360\\
& {Textual entailmen (ConjNLI) } & Macro F1 & .348 & .356 & .368 & .418 & .426 & .441 & .358 & .374 & .405 & .449 & .473 & .512 & \textbf{.524} & .294\\

 \cmidrule(lr){2-17}
& {Multi-choice QA (math \& logic)} & Acc & .450 & .450 & .435 & .445 & .435 & .415 & .525 & .585 & .645 & .605 & .670 & \textbf{.725} & .395 & -\\
 & {Elementary school} & Acc & .535 & .435 & .565 & .590 & .520 & .545 & .580 & .605 & .665 & .660 & .685 & \textbf{.740} & - & -\\
 & Math problems & Math & .209 & .375 & .503 & .194 & .348 & .408 & .412 & .485 & .537 & .458 & .496 & \textbf{.564} & - & -\\
 
\midrule \midrule
  \multirow{2}{*}{\rotatebox{90}{Know}} & 
  {Multi-choice QA (literature)} & Acc & .280 & .295 & .275 & .310 & .305 & .290 & {.440} & .425 & .390 & \textbf{.485} & .470 & .460 & .335 & .250\\

  & {Multi-choice QA (common)} & Acc & .385 & .395 & .445 & .425 & .430 & .430 & .540 & .565 & .595 & .605 & .620 & \textbf{.635} & .250 & .250 \\

 \bottomrule
\end{tabular}}
\caption{Performance of GPT-3.5 and GPT-4 for various datasets compared to the task-specific fine-tuned SOTA models. The highest score for each dataset is in bold.}
\label{tab:classictasks}
\end{table*}


\subsection{Results}
\label{sec:evaluation}

Table \ref{tab:classictasks} displays the performance of ChatGPT models alongside SOTA models across various tasks. The evaluation results are presented for both Persian and English prompts in three modes: zero-shot, one-shot, and three-shots. These tasks are categorized into three subsections: classic, reasoning, and knowledge.

\subsubsection{Classic NLP Tasks}

\paragraph{Classification}
For the \textit{sentiment analysis} task,GPT-4 achieves a peak macro F1 performance of 0.906 when using a three-shot approach with an English prompt. The state-of-the-art (SOTA) model (MT5-base) attains an F1 score of 0.891, surpassing GPT-3.5 but falling short of GPT-4.
Regarding the impact of few-shot examples on ChatGPT models, it is evident that in GPT-4, increasing the number of shots clearly boosts performance. 
However, in GPT-3.5, while few-shot surpasses zero-shot, additional shots do not notably improve results. Intriguingly, with an English prompt in GPT-3.5, the three-shot setting decreases performance, indicating that an increased presence of Persian examples with the English prompt negatively impacts performance.

In the context of \textit{emotion recognition}, a modified version of ParsBERT used in \citet{abaskohi2022persian}, outperforms other models, exhibiting an F1 score of 0.699. In comparison, the highest F1 score attained by ChatGPT models for this task is .621, with a three-shot approach and an English prompt.

In \textit{named entity recognition}, the latest SOTA achievement, boasting an F1-score of 0.988, outperforms other models. GPT-4 achieves the highest F1-score of 0.712 using a three-shot method alongside an English prompt.


\paragraph{Generation}

The performance of \textit{MT} models is shown in Table \ref{tab:classictasks}. 
In English to Persian translation, GPT-4 achieves a notable 8.7 BLEU score, marking the highest when provided with 3 shots and an English prompt. Conversely, in Persian to English translation tasks, SOTA demonstrates superior performance, achieving a BLEU score of 11.7.

\paragraph{Span retrieval}
We use the F1-score metric \citep{rajpurkar-etal-2016-squad} to evaluate \textit{reading comprehension}, which requires finding an answer span in a paragraph for a given question. Table \ref{tab:classictasks} shows that ChatGPT models have lower scores than the SOTA on this task. 
It also shows that one-shot results are better than zero-shot results in both Persian and English prompts by large margins. 
We note that ChatGPT tends to give long answers in zero-shot mode for this task. This lowers the score because the test data has shorter answers, therefore, using one-shot or three-shot modes is helpful.

\subsubsection{Reasoning Tasks}
\label{level3H}

\paragraph{Classification}
We use standard classification metrics, such as F1, accuracy, precision, and recall, to evaluate classification tasks. We calculate these metrics both on average and for each class. We also test the \textit{textual entailment} task with two different datasets: ParsiNLU and ConjNLI.

For the ParsiNLU evaluation, we note that the SOTA F1 score is 0.690 with the fine-tuned MT5-large model. This score is higher than all ChatGPT experiments. 
Comparing GPT-3.5 and GPT-4, we find that GPT-4 achieves higher scores, approaching the SOTA performance. Furthermore, increasing the number of shots enhances results for GPT-4. 
When using Persian prompts with GPT-3.5, performance becomes comparable to random. This suggests that GPT-3.5 struggles to understand task goals in Persian. Analyzing the confusion matrices for experiments with Persian prompts reveals that it often predicts a “Neutral” label. Interestingly, this behavior does not occur with English prompts. The difference implies that the clarity of the Persian prompt may impact the model’s ability to make meaningful inferences between sentences

For the ConjNLI dataset, the highest SOTA F1-score is 0.520, achieved by the fine-tuned xlm-roberta-large model.
Similar to the ParsiNLU dataset, we observe that English prompts yield better results than Persian prompts, but both fall short of the SOTA performance.
Notably, GPT-4 demonstrates noticeable improvement on this dataset as well.

\paragraph{Math \& Logic}

For \textit{multiple-choice QA (math \& logic)},
we employ the accuracy metric as our evaluation criterion. As shown in Table \ref{tab:classictasks}, GPT-4 achieves the best result with an accuracy of 0.725. Both GPT-3.5 and GPT-4 exhibit superior performance compared to SOTA for both Persian and English prompts. In GPT-3.5, introducing additional examples in the prompt does not lead to an improvement. In GPT-4, however, few-shot works better than zero-shot.
One notable challenge during the zero-shot experiments is the model’s tendency to output answers in a different format than the provided answer options. For instance, it might produce ``3!'' instead of ``6'' requiring post-processing adjustments.

For the \textit{elementary school questions} task, since there is no established baseline, we compare the ChatGPT models directly.
Our results indicate that GPT-4 outperforms GPT-3.5, achieving a superior accuracy of 0.740 compared to 0.545. Notably, our analysis reveals significant advancements in few-shot learning with GPT-4. This phenomenon is absent in GPT-3.5.
Furthermore, English prompts consistently yield better results in both models.

We use the ``Math metric’’ \citep{hendrycksmath2021} to evaluate \textit{mathematical problems}. 
The best result achieved by GPT-4 is an accuracy of 0.564. This performance is obtained using an English prompt and a three-shot approach. 
In GPT-3.5 we find that using the Persian prompt in the three-shot setting gives the best performance. Increasing the number of shots in the prompt improves results for both languages and models.

\subsubsection{Knowledge}
We use a multiple-choice QA task to test ChatGPT’s knowledge in two different domains: literature and common knowledge. The main evaluation metric is accuracy.

For literature knowledge, we use the \textbf{multiple-choice QA (literature)} dataset. 
The highest result is achieved by GPT-4 when provided with an English prompt in the zero-shot setting, yielding an accuracy of 0.485.
Interestingly, as the number of shots increases, the results for both models decline.
Notably, GPT-3.5 exhibits notably low performance in this dataset, indicating its lack of knowledge in the literature domain.

For common knowledge, we use the \textbf{multiple-choice QA (common knowledge)} dataset. 
GPT-4 achieves its best accuracy using an English prompt and three shots. Its best result is notably higher than the GPT-3.5's (0.635 vs 0.445), and the SOTA's (0.635 vs 0.250). 
The questions in this domain likely cover diverse topics, making it hard to learn generalizable patterns from the training data. This may explain why a smaller model, like the SOTA here, with less encoded knowledge, does not perform well.

\subsection{Influence of Sample Translation}
\label{sec:sample_translation}

\begin{table*}[h]
    \centering
    \begin{tabular}{l|ccc| c}
    \textbf{Task}  & \multicolumn{3}{c}{\textbf{Translated to EN}} & \textbf{Best-Original}\\
    & zero-shot            & one-shot    & three-shot       &     \\ \toprule
    Sentiment analysis             & .803$_{±.009}$          & .808$_{±.007}$ & \textbf{.811$_{±.005}$} & .804\\
    Multi-choice QA (math \& logic) & .528$_{±.001}$          & .548$_{±.007}$ & \textbf{.567$_{±.008}$} & .450\\
    Multi-choice QA (literature)   & \textbf{.327$_{±.012}$} & .329$_{±.013}$ & .324$_{±.007}$   & .310\\      
    Elementary school questions     & \textbf{.667}$_{±.011}$          & .568$_{±.014}$ & .522$_{±.016}$ & .620 \\
    \bottomrule
    \end{tabular}
    
    \caption{GPT-3.5's performance in four tasks, after translating the samples from Persian to English, in zero- and few-shot scenarios. To compare, we also include the best result of GPT-3.5 using the original Persian samples for each task from Table \ref{tab:classictasks}.}
    \label{tab:demo_translation}
\end{table*}

\begin{table*}[h!]
\centering
\resizebox{12cm}{!}
{\begin{tabular}{llccccccc}
\toprule
 & &\multicolumn{7}{c}{OpenChat}\\
 \cmidrule(lr){4-9}
 & & & \multicolumn{3}{c}{\makecell{Persian Prompt \\ (N-shot)}} &\multicolumn{3}{c}{\makecell{English Prompt\\ (N-shot)}} \\
\cmidrule(lr){4-6}\cmidrule(lr){7-9}
  & Task & & 0 & 1 & 3 & 0 & 1 & 3 \\
  
\midrule
 \multirow{6}{*}{\rotatebox{90} {Classic}} & Sentiment analysis & Macro F1& .460 & .484 & .439 & \textbf{.485} & .466 & .468 \\
& Emotion recognition & Macro F1 & .186 & .327 & .400 & \textbf{.464} & .456 & .454 \\
& NER & Macro F1 & .241 & .603 & \textbf{.606} & .536 & .563 & .588  \\
\cmidrule(lr){2-9}
 & MT ($En \rightarrow Fa$) & Bleu & 5.7 & 6.3 & 6.5 & 5.9 & 6.7 & \textbf{6.8} \\
 & MT ($Fa \rightarrow En$) & Bleu & 9.1 & 9.1 & 9.1 & 9.1 & 9.6 & \textbf{9.6} \\
 \cmidrule(lr){2-9}

& Reading comprehension & F1 &  .506 & .528 & .568 & .595 & .589 & \textbf{.613}\\

\midrule \midrule 

\multirow{5}{*}{\rotatebox{90}{Reasoning}} & {Textual entailment (Parsinlu)} & Macro F1 &  .338 & .468 & .443 & .432 & \textbf{.612} & .554 \\
& {Textual entailmen (ConjNLI) } & Macro F1 & .370 & .415 & .445 & .515 & \textbf{.555} & .555\\

 \cmidrule(lr){2-9}
& {Multi-choice QA (math \& logic)} & Acc & .180 & .260 & \textbf{.300} & .275 & .215 & .245\\
 & {Elementary school} & Acc & .555 & .455 & .520 & \textbf{.585} & .540 & .535\\
 & Math problems & Math & .128 & .229 & \textbf{.241} & .113 & .168 & .214\\
 
\midrule \midrule
  \multirow{2}{*}{\rotatebox{90}{Know}} & 
  {Multi-choice QA (literature)} & Acc & .215 & \textbf{.275} & .24 & .265 & .205 & .265\\

  & {Multi-choice QA (common)} & Acc & .345 & .310 & .300 & .305 & \textbf{.360} & .325\\

 \bottomrule
\end{tabular}}
\caption{Performance of OpenChat-3.5 for various tasks. The best scores are in bold for each row.}
\label{tab:openchat}
\end{table*}

Based on our findings in Table \ref{tab:classictasks}, our investigation reveals an intriguing pattern in GPT-3.5's performance during few-shot scenarios when provided with English prompts: as the number of demonstrations increases, we observe a noticeable decrease in model performance in four tasks. 
Motivated by these results, we delve into the impact of translating samples from Persian to English. By effectively creating an English task for the GPT-3.5 model, we aim to understand how language differences influence performance.
Using Google Translate, we translate the samples either in the prompt or at test time to English. The results are summarized in Table \ref{tab:demo_translation}. We conducted these experiments specifically for the four tasks listed in Table \ref{tab:classictasks} that exhibit a performance drop with more demonstrations.

First, we find that translating the samples to English leads to improved performance. This underscores the significant impact of language differences on GPT-3.5’s behavior. Even automatic translation to English can yield better results than using the original language.

Then, Despite translated demonstrations, the performance continues to deteriorate in the elementary school questions task as more demonstrations are added. For other tasks, especially \textit{sentiment analysis} and \textit{multi-choice QA (math \& logic)}, adding more demonstrations is helpful. 

\subsection{Evaluating OpenChat}

While our initial focus was on ChatGPT models, we later incorporated an open-source multilingual large language model (LLM) with strong reported results in English: OpenChat-3.5 \cite{wang2024openchat}. This model also supports the Persian language.
OpenChat-3.5 is a seven billion parameter open-source model\footnote{\url{https://huggingface.co/openchat/openchat\_3.5}}, which leverages mixed-quality data, combining expert and sub-optimal data without preference labels, achieving strong performance through a novel C-RLFT framework without relying on costly human preference labeling. 
We use this model with their recommended 
the temperature of 0.5, and perform inferences in similar settings as our ChatGPT experiments.

The results are summarized in Table \ref{tab:openchat}. Additionally, Figure \ref{fig:results_overview} provides an overview, allowing us to compare OpenChat-3.5’s best results against those of ChatGPT models.
Accordingly, OpenChat-3.5 falls short compared to GPT-3.5 across most tasks, except for entailment tasks. In tasks such as Named Entity Recognition (NER) and elementary school questions, OpenChat-3.5 performs comparably to GPT-3.5.
Similar to GPT-3.5, OpenChat-3.5 exhibits cases where few-shot performance is worse than zero-shot.

\subsection{Discussion}
\label{sec:discussion}
Here, we summarize and discuss some of the experimental findings of this work. 

We observed that GPT-4 consistently outperforms both GPT-3.5 and OpenChat-3.5 across all our Persian language tasks, demonstrating its robustness and adaptability. While OpenChat-3.5 exhibits marginal improvements in specific datasets, it generally falls short of GPT-3.5’s performance in most tasks. Considering its size and open-source nature, OpenChat-3.5 remains remarkable, and further investigation would enhance our understanding of this model.

A task-specific analysis highlights distinct strengths: SOTA models excel in classic tasks and textual entailment, whereas GPT-4 shines in sentiment analysis. When compared to OpenChat-3.5 and other SOTA models, GPT-4 demonstrates exceptional proficiency in QA reasoning and solving mathematical problems, showcasing unique strengths in these areas.
In knowledge assessment, ChatGPT models exhibit a solid grasp of general knowledge, displaying a broad understanding of common subjects and topics. In more specialized and culture-specific domains, such as Persian literature, GPT-4 clearly outperforms its counterparts.

We have also observed that few-shot learning consistently enhances the performance of GPT-4, resulting in improved results as the number of shots increases. However, this effect is not notably observed in GPT-3.5 and OpenChat-3.5, suggesting that these models behave differently when learning from few-shot samples in prompts.

The choice of language in zero-shot evaluations is crucial. Notably, experiments with English prompts consistently yield better results than those with Persian. Additionally, translating examples into English appears to improve performance in most tasks when testing GPT-3.5. These findings highlight the importance of language selection in test settings and call for deeper investigation into the factors contributing to this difference. For a robust multilingual LLM, limiting its performance to English prompts is not desirable, especially for non-English speakers. Further research in this area could optimize model performance across various languages and prompt languages.

Our findings underscore the complexity of model behavior, emphasizing the critical role of prompt design in optimizing performance across tasks and languages. Additional research is warranted to explore these nuances and enhance model effectiveness across diverse linguistic contexts.

\section{Conclusion}
We evaluated GPT-3.5-turbo, GPT-4, and OpenChat-3.5 on different Persian NLP tasks, grouped into classic, reasoning, and knowledge tasks.  Our evaluation involved testing prompts in both Persian and English, considering zero-shot, one-shot, and three-shot settings to identify optimal prompts for each task. We compared the SOTA task-specific fine-tuned models against LLMs, and presented a comprehensive study.
Our detailed evaluation shows both the strengths and weaknesses of LLMs in different linguistic domains and gives valuable insights for future improvements.
We aim to include other LLMs that support Persian in the future. We also intend to expand our evaluation set to encompass a wider range of tasks, focusing on specific use cases within the Persian-speaking community.



\nocite{*}
\section{Bibliographical References}
\label{lreference}

\bibliographystyle{lrec-coling2024-natbib}
\renewcommand\refname{}
\bibliography{lrec-coling2024-example}


\appendix

\section{SOTA models}
\label{appen:SOTA}

\begin{table*}[t]
\centering
\footnotesize
\begin{tabular}{ll}
\toprule
\textbf{Task} & \textbf{Models} \\

\toprule
\multirow{3}{*}{Sentiment Classification} & \href{https://huggingface.co/persiannlp/mt5-small-parsinlu-sentiment-analysis}{mt5-small-parsinlu-sentiment-analysis} \\
 & \textbf{\href{https://huggingface.co/persiannlp/mt5-base-parsinlu-sentiment-analysis}{mt5-base-parsinlu-sentiment-analysis}} \\
 & \href{https://huggingface.co/persiannlp/mt5-large-parsinlu-sentiment-analysis}{mt5-large-parsinlu-sentiment-analysis} \\
 
\toprule
\multirow{5}{*}{Textual Entailment (ParsiNLU)}  &  \href{https://huggingface.co/persiannlp/wikibert-base-parsinlu-entailment}{wikibert-base-parsinlu-entailment}\\
 & \href{https://huggingface.co/persiannlp/mt5-base-parsinlu-snli-entailment}{mt5-base-parsinlu-snli-entailment}\\
 & \textbf{\href{https://huggingface.co/persiannlp/mt5-large-parsinlu-snli-entailment}{mt5-large-parsinlu-snli-entailment}}\\
 & \href{https://huggingface.co/persiannlp/parsbert-base-parsinlu-entailment}{parsbert-base-parsinlu-entailment}\\
 & \href{https://huggingface.co/persiannlp/mbert-base-parsinlu-entailment}{mbert-base-parsinlu-entailment}  \\

\toprule
\multirow{2}{*}{Textual Entailment (ConjNLI)}  & \textbf{\href{https://huggingface.co/FacebookAI/xlm-roberta-large}{xlm-roberta-large}}\\
 & \href{https://huggingface.co/google-bert/bert-base-multilingual-cased}{bert-base-multilingual-cased}\\
 & \href{https://huggingface.co/google/mt5-large}{mt5-large}\\

\toprule
\multirow{1}{*}{Named Entity Recognition}  & \textbf{\href{https://huggingface.co/HooshvareLab/bert-fa-base-uncased-ner-arman}{Bert-fa-base-uncased-ner-arman}}\\

\toprule
\multirow{6}{*}{Multiple-Choice QA}  &  \textbf{\href{https://huggingface.co/persiannlp/mt5-small-parsinlu-multiple-choice}{mt5-small-parsinlu-multiple-choice}}(best on literature)\\
 & \href{https://huggingface.co/persiannlp/mt5-base-parsinlu-multiple-choice}{mt5-base-parsinlu-multiple-choice}\\
 & \textbf{\href{https://huggingface.co/persiannlp/mt5-large-parsinlu-multiple-choice}{mt5-large-parsinlu-multiple-choice}}(best on math\&logic)\\
 & \href{https://huggingface.co/persiannlp/mt5-small-parsinlu-arc-comqa-obqa-multiple-choice}{mt5-small-parsinlu-arc-comqa-obqa-multiple-choice}\\
 & \href{https://huggingface.co/persiannlp/mt5-base-parsinlu-arc-comqa-obqa-multiple-choice}{mt5-base-parsinlu-arc-comqa-obqa-multiple-choice} \\
 & \textbf{\href{https://huggingface.co/persiannlp/mt5-large-parsinlu-arc-comqa-obqa-multiple-choice}{mt5-large-parsinlu-arc-comqa-obqa-multiple-choice}}(best on com-know)\\

\toprule
\multirow{3}{*}{Reading Comprehension} &  \href{https://huggingface.co/persiannlp/mt5-small-parsinlu-squad-reading-comprehension}{mt5-small-parsinlu-squad-reading-comprehension} \\
 & \href{https://huggingface.co/persiannlp/mt5-base-parsinlu-squad-reading-comprehension}{mt5-base-parsinlu-squad-reading-comprehension}\\
 & \textbf{\href{https://huggingface.co/persiannlp/mt5-large-parsinlu-squad-reading-comprehension}{mt5-large-parsinlu-squad-reading-comprehension}}\\

\toprule
\multirow{4}{*}{Emotion Classification} & \href{https://huggingface.co/Toshifumi/distilbert-base-multilingual-cased-finetuned-emotion}{distilbert-base-multilingual-cased-finetuned-emotion}  \\
 & \href{https://huggingface.co/MilaNLProc/xlm-emo-t}{xlm-emo-t}\\
 & \textbf{\href{https://github.com/AmirAbaskohi/Persian-Emotion-Detection-using-ParsBERT-and-Imbalanced-Data-Handling-Approaches}{ParsBERT-and-Imbalanced-Data-Handling-Approaches}}\\
 & \href{https://huggingface.co/Toshifumi/bert-base-multilingual-cased-finetuned-emotion}{bert-base-multilingual-cased-finetuned-emotion}\\

\toprule
Translation  & \href{https://huggingface.co/persiannlp/mt5-small-parsinlu-opus-translation_fa_en}{mt5-small-parsinlu-opus-translation\_fa\_en} \\
 & \href{https://huggingface.co/persiannlp/mt5-base-parsinlu-opus-translation_fa_en}{mt5-base-parsinlu-opus-translation\_fa\_en}\\
 & \textbf{\href{https://huggingface.co/persiannlp/mt5-large-parsinlu-opus-translation_fa_en}{mt5-large-parsinlu-opus-translation\_fa\_en}} (Persian to English)\\
 & \href{https://huggingface.co/persiannlp/mt5-small-parsinlu-translation_en_fa}{mt5-small-parsinlu-translation\_en\_fa}\\
 & \href{https://huggingface.co/persiannlp/mt5-base-parsinlu-translation_en_fa}{mt5-base-parsinlu-translation\_en\_fa}\\
 & \textbf{\href{https://huggingface.co/persiannlp/mt5-large-parsinlu-translation_en_fa}{mt5-large-parsinlu-translation\_en\_fa}} (English to Persian)\\

\bottomrule
\end{tabular}
\caption{Evaluated tasks and fine-tuned models. Highlighted in \textbf{bold} are models showcasing superior performance.}
\label{tab:SOTA_Models}
\end{table*}

Before the emergence of large language models, to achieve the best accuracy on a dataset, one way was to fine-tune pre-trained models on the training subset of a dataset for a specific task, but fine-tuning for large language models is not efficient and one of the outstanding capabilities of LLMs is generalization and lack of need for task-specific fine-tuning. Therefore, we have selected a series of the best language models that are specifically fine-tuned on the training subset of our targeted datasets, we have called them SOTA models. Then we use them to compare with ChatGPT.
Table~\ref{tab:SOTA_Models} lists the names of the fine-tuned models available for our targeted datasets for each task. All models are available on HuggingFace or Github.
In sentiment classification, Persian textual entailment (ParsiNLU), multiple-choice QA, reading comprehension, and translation tasks, SOTA models are fine-tuned on the training subsets of the ParsiNLU datasets. In the same way, the chosen model for NER is fine-tuned using the ArmanNER training dataset. Emotion classification involves the testing of certain multilingual models fine-tuned specifically for this purpose. When dealing with textual entailment of the English dataset (ConjNLI), popular multilingual models suffice for the task.

\section{Prompts}
\label{sec:appendix:prompts}
In the below, we provide an example prompt for each task.

\subsection{Sentiment Analysis}

\hrule\vspace{4pt}
The below sentence is a person's review. The review is in Persian. Identify the sentiment or polarity associated with it. \\
Possible answers are: POSITIVE, NEUTRAL, NEGATIVE, OTHER.\\
Use OTHER when the sentence does not include any specific sense, or has mixed or borderline senses.

\noindent\dotfill Just for K-shots (K>0) \noindent\dotfill \\
Examples: \\
Review: \{ExampleInput\} \\
Sentiment: \{ExampleLabel\} \\
\vdots

\noindent\dotfill \\
Review: \{TestInput\} \\
Sentiment:
\vspace{4pt}\hrule

\subsection{Textual entailment}

\hrule\vspace{4pt}
Natural Language Inference: Read the following premise and hypothesis carefully and determine the relationship between them.
Choose one of the three categories below that best describes their relationship:
- entailment: The meaning of the hypothesis is logically inferred or derived from the premise.
- contradiction: The meaning of the hypothesis contradicts or conflicts with the premise.
- neutral: There is no clear logical relationship between the premise and hypothesis.
Note: The premise and hypothesis are in Persian.
example pattern:
<premise><sep><hypothesis>
<category>:
    entailment or contradiction or neutral
    
\noindent\dotfill Just for K-shots (K>0) \noindent\dotfill \\
Examples: \\
Please select the appropriate category for the given example: \{ExampleInput\} \\
category: \{ExampleLabel\} \\
\vdots

\noindent\dotfill \\
Please select the appropriate category for the given example: \{TestInput\} \\
category:
\vspace{4pt}\hrule

\subsection{Reading Comprehension}

\hrule\vspace{4pt}
In this task, you will be shown a Persian passage and question. You need to write an answer for the question. Try to keep your answers as short as possible.

\noindent\dotfill Just for K-shots (K>0) \noindent\dotfill \\
Examples: \\
Review: \{ExampleInput\} \\
Answer: \{ExampleAnswer\} \\
\vdots

\noindent\dotfill \\
Review: \{TestInput\} \\
Answer:
\vspace{4pt}\hrule

\subsection{Multiple Choice}

\hrule\vspace{4pt}
In this task, you will be presented with a multiple-choice question in Persian, and you should answer the question based on your knowledge. choose the answer from the given candidates.

\noindent\dotfill Just for K-shots (K>0) \noindent\dotfill \\
Examples: \\
Review: \{ExampleInput\} \\
Answer: \{ExampleAnswer\} \\
\vdots

\noindent\dotfill \\
Review: \{TestInput\} \\
Answer:
\vspace{4pt}\hrule

\subsection{NER}

\hrule\vspace{4pt}
Task Description: \\
You need to label a given Persian token list with named entity labels. \\
Named Entity Labels: \\
PER (person) \\
LOC (location) \\
 ORG (organization) \\
 Product (Product)  \\
 Event (Event)  \\
 Facility (Facility) \\
 Output Format: \\
Your output format should be a list of tuples, where each tuple consists of a word from the input text and its corresponding named entity label. For words which are not part of any named entity, you should return 'O'. 

\noindent\dotfill Just for K-shots (K>0) \noindent\dotfill \\
Examples: \\
Input: \{InputSentence\} \\
Label: \{LabelSentence\}
\vdots

\noindent\dotfill \\
Input: \{TestInput\} \\
Label: 
\vspace{4pt}\hrule

\subsection{Translation}

\hrule\vspace{4pt}
In this task, which is Machine Translation (MT), you will  be presented with a sentence in \{source language\}.
You should translate it to \{target language\} in the most appropriate way. \\
\{source language\} to \{target language\}: \{sentence\}

\noindent\dotfill Just for K-shots (K>0) \noindent\dotfill \\
Examples: \\
Translate it: \{SourceLanguageText\} -> \{TargetLanguageText\} \\
\vdots

\noindent\dotfill \\
Translate it: \{TestInput\} \\
\vspace{4pt}\hrule

\subsection{Mathematical Problems}

\hrule\vspace{4pt}
You are specialized in mathematics. I would give you a problem and I want you to provide a clear and step by step solution for that problem and also give me the final result. Also, make sure that your anwer is it persian. I would give you the problem and I expect you to answer in the format below:

\noindent\dotfill Just for K-shots (K>0) \noindent\dotfill \\
Examples: \\
problem \\
\{ExampleInput\} \\
/problem\\
solution\\
\{ExampleSolution\} \\
/solution\\
answer\\
\{ExampleAnswer\} \\
/answer
\vdots
\noindent\dotfill \\
problem \\
\{TestInput\} \\
/problem\\

\vspace{4pt}\hrule

\subsection{Emotion Recognition}

\hrule\vspace{4pt}
The below sentence is a person's tweet. The tweet is in Persian. Identify the emotion or sentiment associated with it. \
Possible answers are: OTHER, ANGRY, SAD, FEAR, SURPRISE, HAPPY, HATE. \
Use OTHER when the sentence does not include any specific emotion, has mixed emotion, or has emotion other than the classes mentioned. \

\noindent\dotfill Just for K-shots (K>0) \noindent\dotfill \\
Examples: \\
Tweet: \{ExampleInput\} \\
Emotion: \{ExampleLabel\} \\
\vdots

\noindent\dotfill \\
Tweet: \{TestInput\} \\
Emotion:
\vspace{4pt}\hrule

\subsection{Elementary School Questions}

\hrule\vspace{4pt}
  In this task, you will be shown a Persian elementary math question. You need to write an answer for the question.
  the question is a two choice question.
  The answer is either 'A' or 'B'.
  So you should answer only in one letter, 'A' or 'B'.\\
\noindent\dotfill Just for K-shots (K>0) \noindent\dotfill \\
Examples: \\
iquestion: \{ExampleInput\} \\
answer: \{ExampleAnswer\} \\
\vdots

\noindent\dotfill \\
question: \{TestInput\} \\
answer:
\vspace{4pt}\hrule

\section{Authors Contributions}
\label{sec:appendix:contributions}
\textbf{Amirhossein Abaskohi} assessed GPT-4's performance across all tasks and specifically examined GPT-3.5's capabilities in emotion recognition and sample translation effects. He also contributed to creating Math problem benchmarks. Additionally, he helped with writing the paper.
\textbf{Sara Baruni} was the primary writer of the paper and also helped with coordinating the tasks among the team.
\textbf{Mostafa Masoudi}
examined GPT-3.5 and OpenChat for Textual entailment (Parsinlu). Additionally, he helped with writing the paper.
\textbf{Nesa Abbasi} and  \textbf{Mohammad Hadi Babalou} examined GPT-3.5 and OpenChat for sentiment analysis. 
\textbf{Ali Edalat}
examined GPT-3.5 and OpenChat for NER. He also helped with examining the OpenChat for all other tasks. 
\textbf{sepehr kamahi}
examined GPT-3.5 and OpenChat in Elementary school questions and created a benchmark for Elementary school task.
\textbf{Samin Mahdizadeh Sani}
examined GPT-3.5 and OpenChat for all multiple-choice QA and reading comprehension tasks.
\textbf{Nikoo Naghavian} translated the 200 samples of ConjNLI to Persian and examined GPT-3.5 and OpenChat for Textual entailment (ConjNLI). 
\textbf{Danial Namazifard}
examined GPT-3.5 and OpenChat for MT.
\textbf{Pouya Sadeghi}
compiled the samples of Math problems, and 
examined GPT-3.5 and OpenChat for that. He is also responsible for the GitHub repository.
\textbf{Yadollah Yaghoobzadeh} was the initiator and supervisor of this project. He helped with the planning, writing, and coordinating the team.

\end{document}